\documentclass[11pt]{article}
\usepackage{acl}

\usepackage{times}
\usepackage{latexsym}
\usepackage[T1]{fontenc}
\usepackage[utf8]{inputenc}
\usepackage{microtype}
\usepackage{inconsolata}
\usepackage{graphicx}
\usepackage{booktabs}
\usepackage{multirow}
\usepackage{amsmath}
\usepackage{array}
\usepackage{xcolor}
\usepackage{colortbl}
\usepackage{pgfplots}
\pgfplotsset{compat=1.17}
\usepackage{fancyvrb}

\definecolor{pbTealDark}{HTML}{005261}
\definecolor{pbTeal}    {HTML}{008290}
\definecolor{pbCyan}    {HTML}{13E7FE}
\definecolor{pbMaroon}  {HTML}{700000}
\definecolor{pbRed}     {HTML}{EF1D32}
\definecolor{pbRedLight}{HTML}{FF4E4E}
\definecolor{pbAmber}   {HTML}{EAA10F}
\definecolor{pbGold}    {HTML}{FFC550}

\definecolor{goodcell}{HTML}{D9EFF1}   
\definecolor{okaycell}{HTML}{FFF1CE}   
\definecolor{badcell} {HTML}{FBDCDC}   
\definecolor{pbBoxBg}  {HTML}{F5F9F9}   
\definecolor{pbBoxRule}{HTML}{C9DDDD}   

\newsavebox{\pbpromptbox}
\newenvironment{promptbox}[1][]{%
	\par\medskip
	\def\pbtitle{#1}%
	\ifx\pbtitle\empty\else
	\noindent{\footnotesize\textcolor{pbTealDark}{\textbf{#1}}}\par\nobreak\smallskip
	\fi
	\noindent
	\setlength{\fboxrule}{0.6pt}\setlength{\fboxsep}{6pt}%
	\begin{lrbox}{\pbpromptbox}%
	\begin{minipage}{\dimexpr\columnwidth-2\fboxsep-2\fboxrule\relax}%
	\scriptsize\ttfamily
}{%
	\end{minipage}\end{lrbox}%
	\fcolorbox{pbBoxRule}{pbBoxBg}{\usebox{\pbpromptbox}}\par\medskip
}

\newcommand{\na}{\textcolor{pbRed}{$\ast$}}  
\newcommand{\best}[1]{\textbf{#1}}
\newcommand{\vazn}{\emph{vazn}}
\newcommand{\sabk}{\emph{sabk}}

\title{PARSA-Bench: A Comprehensive Persian Audio-Language Model Benchmark}
\author{
	\textbf{Mohammad Javad Ranjbar Kalahroodi$^{1,*}$}
	\quad
	\textbf{Mohammad Amini$^{1,*}$}
	\\
	\textbf{Parmis Bathayan$^{1}$}
	\quad
	\textbf{Heshaam Faili$^{1}$}
	\quad
	\textbf{Azadeh Shakery$^{1,2}$}
	\\[2mm]
	\small $^{1}$School of Electrical and Computer Engineering, University of Tehran, Iran
	\\
	\small $^{2}$Institute for Research in Fundamental Sciences (IPM), Tehran, Iran
	\\
	\small \texttt{\{MohammadJRanjbar,parmis.bathayan,hfaili,shakery\}@ut.ac.ir}
	\\
	\small $^{*}$Equal contribution
}

\begin{document}
	\maketitle
	
	\begin{abstract}
		Persian poses unique audio understanding challenges through its classical poetry, traditional music, and pervasive code-switching, none of which is captured by existing benchmarks. We introduce \textbf{PARSA-Bench} (\textbf{P}ersian \textbf{A}udio \textbf{R}easoning and \textbf{S}peech \textbf{A}ssessment Benchmark), the first dedicated benchmark for evaluating LALMs on Persian language and culture. It covers 16 tasks, ten of them new, spanning speech understanding, paralinguistic analysis, and culturally grounded audio reasoning. Across most tasks, text-only baselines outperform their audio counterparts, so audio understanding rather than language knowledge remains the main limitation, and supplying the transcript alongside the audio lifts weak models to near their text-only level. The consistent exception is Persian poetry, where prosody carries information the written form cannot: audio beats text on both poetry tasks, and metre detection shows the first signs of being learnable only at the largest model scale. The benchmark is publicly available on Hugging Face.\footnote{\url{https://huggingface.co/datasets/MohammadJRanjbar/PARSA-Bench}}
		
	\end{abstract}
	
	\begin{figure}[t]
		\centering
		\includegraphics[width=\columnwidth]{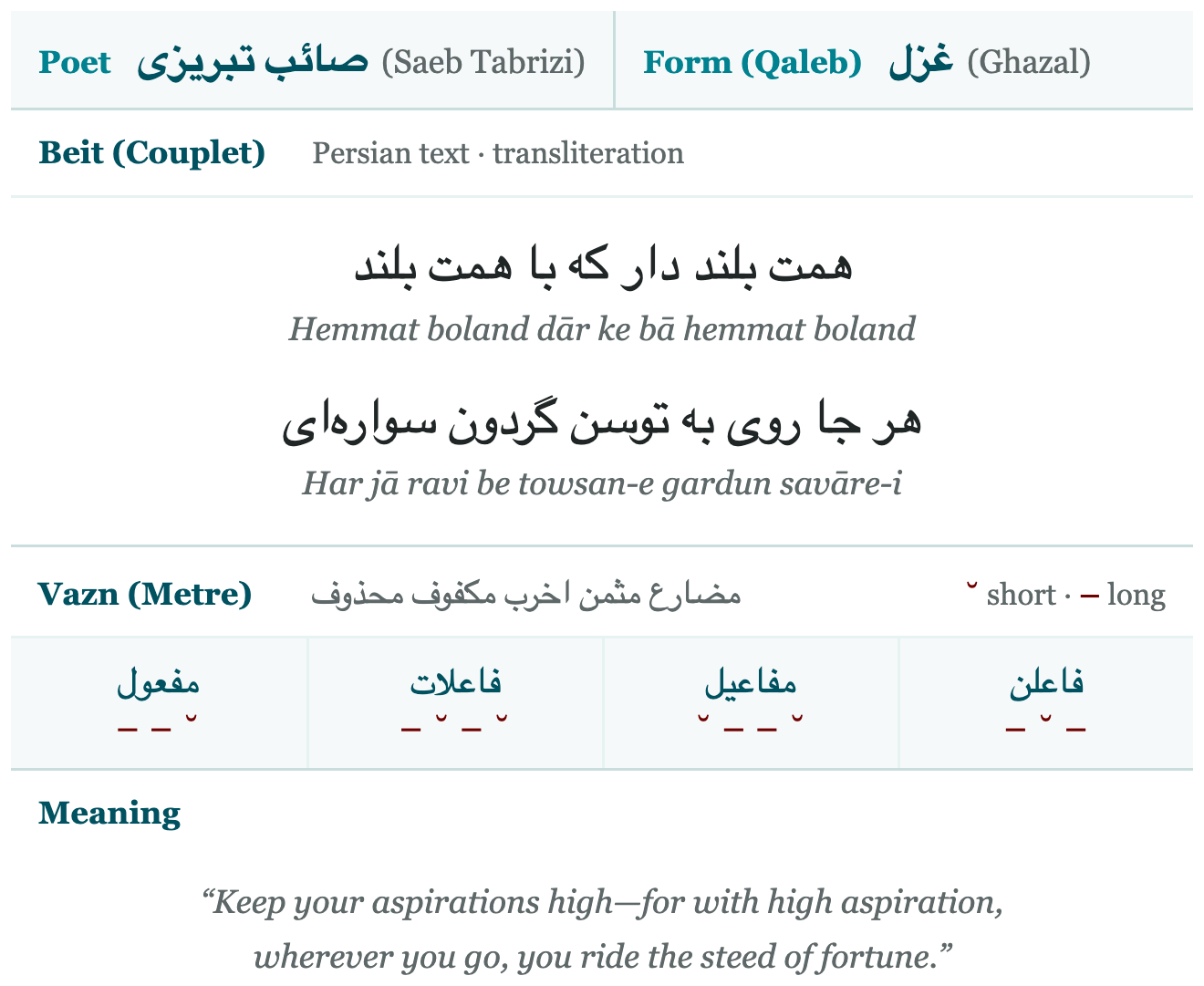}
		\caption{An annotated example from PARSA-Bench. The couplet is labelled with
			its \vazn. Because short vowels are omitted in Persian script, the metre
			cannot be recovered from the written form; the recitation is the primary
			carrier of metrical information.}
		\label{fig:poem}
	\end{figure}
	
	\begin{figure}[t]
		\centering
		\includegraphics[width=\columnwidth]{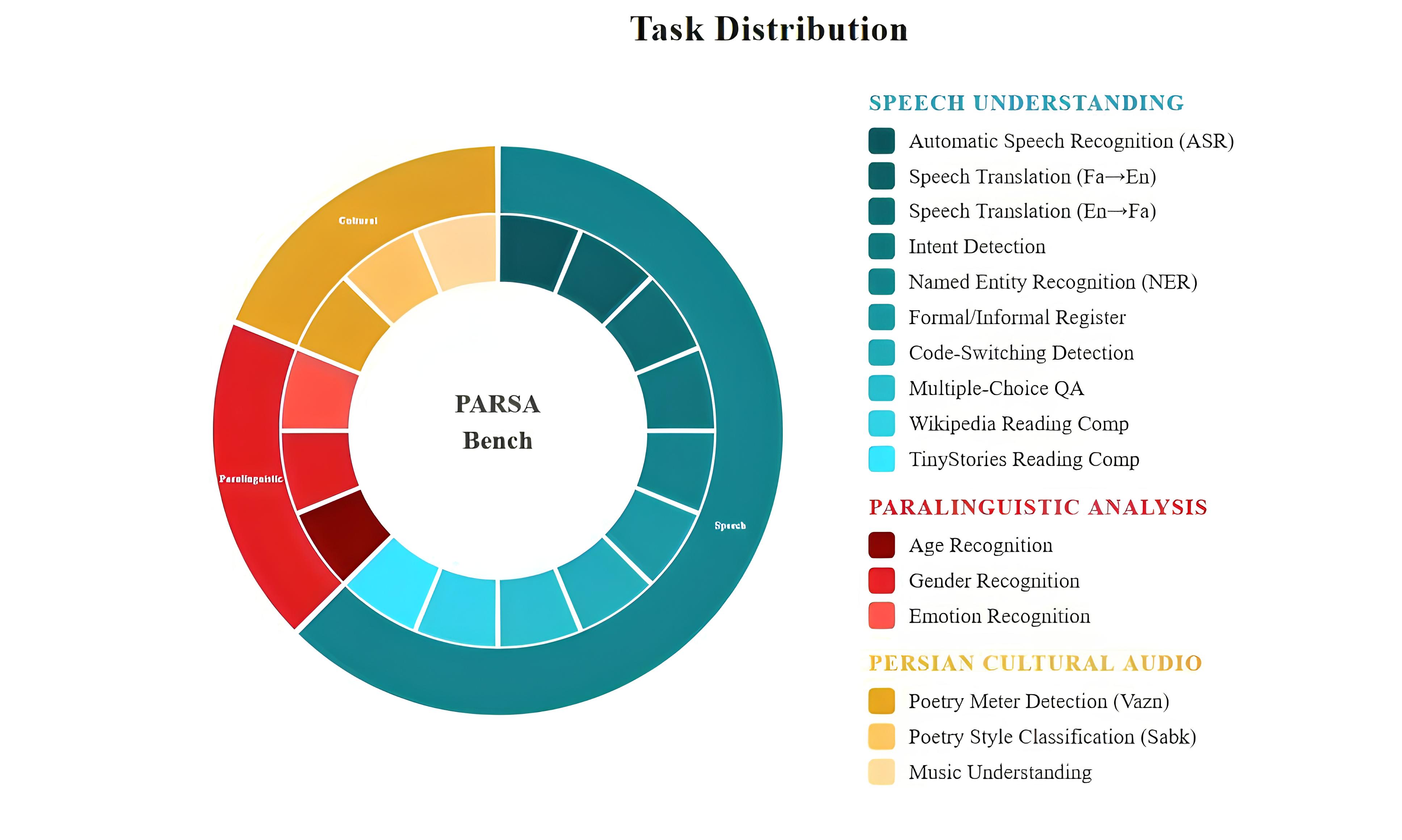}
		\caption{PARSA-Bench spans three evaluation dimensions: Speech
			Understanding, Paralinguistic Analysis and Persian Cultural Audio
			Understanding.}
		\label{fig:overview}
	\end{figure}
	
	\section{Introduction}
	\label{sec:introduction}
	
	Large language models achieve strong performance across textual tasks
	\citep{Brown2020, OpenAI2023GPT4, Touvron2023LLaMA}, but spoken language carries
	information that text cannot represent: tone, prosody, emotion and ambient
	context. Transcribing audio before reasoning discards exactly the signal that
	makes speech rich. Large audio-language models (LALMs) address this by
	processing audio end-to-end; models such as Qwen-Audio \citep{Chu2023QwenAudio}
	and SALMONN \citep{Tang2024SALMONN} demonstrate broad capability across speech,
	environmental sound and music. Their development, however, has centred on
	English and Western cultural content.
	
	Persian, spoken by over 100 million people, is a demanding test case for
	culturally grounded audio understanding. Persian classical poetry is organised
	by quantitative metrical patterns (\vazn) and stylistic traditions (\sabk),
	and continues to be transmitted through oral recitation as an active cultural practice. As Figure~\ref{fig:poem} illustrates, identifying \vazn{} from audio requires perceiving prosodic rhythm that is absent from the transcript. Because short vowels are omitted in standard Persian orthography, metre is not recoverable from the surface text alone, and while a native reader restores the missing vowels from lexical and grammatical knowledge, that information is not encoded in the written form itself. Persian traditional music is organised around the
	Dastgah modal framework, a system absent from Western corpora. Persian--English
	code-switching is common in contemporary Persian media and everyday speech.
	None of these phenomena is covered by existing audio benchmarks: AIR-Bench
	\citep{Qian2024AIRBench} and AudioBench \citep{Wang2024AudioBench} provide broad
	English-centric evaluation, but they offer no mechanism for assessing the unique
	linguistic and cultural challenges that Persian audio poses to current models.
	No dedicated benchmark exists for this purpose.
	
	The challenge of Persian audio understanding goes beyond data scarcity. Persian
	has limited speech training data; its cultural knowledge cannot be obtained by
	translating English resources; and existing evaluation frameworks were not
	designed with such languages in mind. Addressing this requires tasks built from
	the ground up rather than adapted from English templates.
	
	To fill this gap, we introduce \textbf{PARSA-Bench}, a large-scale benchmark covering
	16 tasks covering speech understanding, paralinguistic analysis, and culturally grounded audio understanding, as shown in Figure~\ref{fig:overview}. We
	evaluate eight state-of-the-art LALMs in zero-shot and extended prompting configurations,
	and include text-only baselines to precisely isolate audio processing failures from
	failures of linguistic competence. Our experiments reveal that the audio-text performance
	gap is large and consistent across nearly all tasks, confirming that audio processing, not language
	understanding, is the primary bottleneck. As shown in Table~\ref{tab:main}, culturally-grounded tasks expose a qualitatively
	distinct failure mode in Persian poetry meter detection: even the largest models perform near random chance. Although poetic meter may appear in text pretraining data, accurately identifying vazn requires culturally grounded and prosodic understanding that current models struggle to acquire in both text-only and audio settings.
	
	\section{Related Work}
	\label{sec:related}
	
	\paragraph{Large audio-language models.}
	Early multimodal audio work targeted transcription, captioning and retrieval
	\citep{li2024whisma, peng2023reproducing, wu2023large, elizalde2023clap} but was
	limited on reasoning-centric tasks. Unified audio-language architectures
	\citep{ghosh2024gama, ghosh2025audio} and large audio reasoning models such as
	Audio-Reasoner \citep{xie2025audio} and SoundMind \citep{diao2025soundmind}
	extend this to stepwise reasoning. Evaluation has not kept pace, particularly
	across linguistically and culturally diverse settings.
	
	\paragraph{Audio understanding benchmarks.}
	MMAU \citep{sakshi2024mmau} provides large-scale QA over speech, sound and
	music; MMAR \citep{ma2025mmar} adds hierarchical reasoning. AudioBench
	\citep{Wang2024AudioBench} aggregates many datasets, MuChoMusic
	\citep{weck2024muchomusic} targets music, MMSU \citep{wang2506mmsu} covers
	dozens of spoken-language skills, and Dynamic-SUPERB \citep{huang2024dynamic}
	spans over 180 instruction-tuned tasks. All evaluate audio understanding in
	isolation from non-Western linguistic and cultural context. MMAU-Pro
	\citep{kumar2025mmau} incorporates music from eight regions and exposes a clear
	training-data bias, strongest on Western and Chinese music and weakest on
	Indian, Latin American and Middle Eastern traditions, but the linguistic
	dimension remains more underexplored still. No existing benchmark systematically
	evaluates LALMs on non-English speech understanding, paralinguistic analysis and
	culturally specific audio content: the gap PARSA-Bench addresses.
	
	\begin{table*}[t]
		\centering
		\footnotesize
		\setlength{\tabcolsep}{4pt}
		\begin{tabular}{@{}lrrl@{}}
			\toprule
			\textbf{Task} & \textbf{Eval.} & \textbf{Classes} & \textbf{Data source} \\
			\midrule
			\multicolumn{4}{@{}l}{\textit{\textbf{Speech Understanding}}}\\
			Automatic Speech Recognition & 500 & -- & Common Voice, ParsVoice \\
			Speech Translation (Fa$\rightarrow$En) & 500 & --  & CoVoST2 \citep{Wang2021CoVoST2} \\
			Speech Translation (En$\rightarrow$Fa) & 499 & --  & CoVoST2 \citep{Wang2021CoVoST2} \\
			Intent Detection$^\dagger$             & 500 & 56  & MASSIVE \citep{FitzGerald2023MASSIVE} + TTS \\
			Named Entity Recognition$^\dagger$     & 497 & 42  & MASSIVE \citep{FitzGerald2023MASSIVE} + TTS \\
			Formal/Informal Register$^\dagger$ & 496 & 2 & Mana-TTS \citep{ManaTTS}, GPTInformal-Fa \\
			Code-Switching Detection$^\dagger$     & 500 & 2   & Common Voice + YouTube \\
			Multiple-Choice QA$^\dagger$ & 500 & 4 & ParsiNLU \citep{Khashabi2021ParsiNLU} + TTS \\
			Reading Comp.\ (Wikipedia)$^\dagger$ & 500 & 4 & ParsiNLU \citep{Khashabi2021ParsiNLU} + TTS \\
			Reading Comp.\ (TinyStories)$^\dagger$ & 500 & 4   & TinyStories \citep{eldan2023tinystoriessmalllanguagemodels} \\
			\midrule
			\multicolumn{4}{@{}l}{\textit{\textbf{Paralinguistic Analysis}}}\\
			Age Recognition     & 480 & 6 & Common Voice \citep{Ardila2019CommonVoice} \\
			Gender Recognition  & 500 & 2 & Common Voice \citep{Ardila2019CommonVoice} \\
			Emotion Recognition & 388 & 6 & SHEMO \citep{Nezami2019SHEMO} \\
			\midrule
			\multicolumn{4}{@{}l}{\textit{\textbf{Persian Cultural Audio Understanding}}}\\
			Poetry Metre (\vazn)$^\dagger$  & 347 & 10 & Ganjoor \citep{Ganjoor} \\
			Poetry Style (\sabk)$^\dagger$  & 442 & 4 & Ganjoor \citep{Ganjoor} \\
			Music Understanding$^\dagger$ & 496 & 4 & Persian Music Dataset \citep{esfangereh2025persian} \\
			\midrule
			\textbf{Total: 16 tasks (10 new)} & \textbf{7{,}645} & & \\
			\bottomrule
		\end{tabular}
		\caption{PARSA-Bench tasks. $\dagger$ marks tasks newly introduced for Persian.}
		\label{tab:overview}
	\end{table*}
	
	\section{Benchmark Construction}
	\label{sec:construction}
	
	PARSA-Bench evaluates LALMs along three dimensions, with 7{,}645 evaluated
	samples in total.
	Table~\ref{tab:overview} lists all 16 tasks with the number of items actually
	scored, the size of the label set and the data source.
	Table~\ref{tab:durations} in Appendix~\ref{app:duration} reports audio-duration
	statistics.

	Unless stated otherwise, evaluation subsets were drawn from a larger pool by
	stratified random sampling and are class-balanced wherever the pool permits.
	Emotion contains fewer usable \emph{fear} examples than the other classes,
	and the two poetry tasks preserve the naturally occurring class
	distributions, because balancing them would discard most of the available
	recordings; the exact counts per task are in Table~\ref{tab:overview} and
	Appendix~\ref{app:duration}.
	
	\subsection{Speech Understanding}
	
	\paragraph{ASR and translation.}
	ASR samples from Common Voice \citep{Ardila2019CommonVoice} and ParsVoice
	\citep{kalahroodi2025parsvoicelargescalemultispeakerpersian}, filtered to
	utterances of at least three transcript words. The result spans diverse speaker
	demographics and recording conditions within read speech; spontaneous
	conversational ASR is out of scope.
	Bidirectional Persian--English translation pairs come from CoVoST2
	\citep{Wang2021CoVoST2}.
	
	\paragraph{Intent detection and named entity recognition.}
	Both use MASSIVE \citep{FitzGerald2023MASSIVE}, which provides Persian intent
	and entity annotations. Because MASSIVE is text-only, audio was synthesised with
	the ParsVoice TTS system
	\citep{kalahroodi2025parsvoicelargescalemultispeakerpersian}, using speaker
	prompts drawn from Common Voice as reference voices so that samples vary in
	speaker identity and prosody rather than being read in one synthetic voice.
	Following AudioBench \citep{Wang2024AudioBench}, which finds high-quality TTS an
	acceptable proxy for natural speech in evaluation, a random subset of the
	synthesised audio was manually inspected for synthesis failures such as
	truncation or wrong-language output, and no unusable examples were found.
	
	\paragraph{Formal/informal register.}
	Persian distinguishes registers that carry pragmatic meaning beyond lexical
	content. The formal half is drawn from Mana-TTS \citep{ManaTTS} and the informal
	half from the GPTInformal-Persian collection.
	
	\paragraph{Code-switching detection.}
	We define code-switching at the word level: an utterance is positive if
	at least one English word is embedded in an otherwise Persian utterance.
	Conventionalised borrowings that have entered standard Persian lexicographically
	are not counted as switches, whereas productive insertions such as an English
	technical term inside a Persian clause are. Audio
	comes from Persian YouTube channels and Common Voice. Ground truth is derived
	from transcript metadata and manually verified.
	
	\paragraph{Reading comprehension and MCQA.}
	From ParsiNLU \citep{Khashabi2021ParsiNLU} we build a multiple-choice QA task
	(mathematics and logic, literature, common knowledge) and a
	Wikipedia-passage reading-comprehension task. To reduce overlap with pretraining
	data we add a second reading-comprehension set built on Persian TinyStories
	\citep{eldan2023tinystoriessmalllanguagemodels}, whose passages are novel by
	construction. The MCQA options are ParsiNLU's own answer candidates, used
	unchanged. For both reading-comprehension sets the four options were generated
	with GPT-4o-mini, which was also asked to randomise the position of the correct
	answer; the generation prompts and sampling temperatures are in
	Appendix~\ref{app:prompts}. Generated items were filtered so that no option
	duplicates the gold answer, and a subset was reviewed by a native speaker before
	release.
	
	\subsection{Paralinguistic Analysis}
	
	Age and gender use self-reported Common Voice metadata
	\citep{Ardila2019CommonVoice}; age uses the six Common Voice brackets. Emotion
	uses SHEMO \citep{Nezami2019SHEMO}. Per-task label distributions are reported in
	Appendix~\ref{app:duration}.
	
	\subsection{Persian Cultural Audio Understanding}
	
	\paragraph{Poetry.}
	We crawled the Ganjoor digital library \citep{Ganjoor}, which hosts recitations
	by multiple reciters. For metre we selected the most frequent \vazn{} classes
	and extracted the first two \emph{beits} (couplets), enough to establish the
	pattern. Metre is presented as a ten-option multiple choice over the ten most
	frequent metres. Style uses four canonical \sabk{} categories
	(\emph{ghazal/qasideh/qat'eh}, \emph{masnavi}, \emph{ruba'i}, \emph{dobeyti})
	as four options, with four-\emph{beit} excerpts so that structurally similar
	styles remain distinguishable.
	
	\paragraph{Music.}
	Persian classical music is organised around the Dastgah modal system. From the
	Persian music dataset \citep{esfangereh2025persian} we build one four-option QA
	task pooling seven question types: Dastgah identification, instrument
	identification, instrument presence, instrument count, ensemble size, polyphony
	and tempo. Questions were generated from structured templates and paraphrased
	with GPT-4o-mini for linguistic diversity
	(prompt in Appendix~\ref{app:prompts}).
	
	\section{Evaluation Protocol}
	\label{sec:protocol}
	
	This section describes how models were queried and scored.
	Appendix~\ref{app:prompts} reproduces every prompt verbatim.
	
	\subsection{Prompting and decoding}
	
	All prompts are issued in English, following established practice in
	multilingual LALM evaluation \citep{Wang2024AudioBench, Qian2024AIRBench}, since
	instruction-following is stronger in English regardless of the target language;
	the task content (passages, questions, options) remains in Persian. Every prompt
	has the same three-part shape: a role sentence, the task instruction with the
	explicit label set or option list, and a hard output-format constraint.
	One representative example (the rest are in
	Appendix~\ref{app:prompts}):
	
	\begin{promptbox}
		You are an expert Persian speech recognition system. Listen to the Persian audio
		and transcribe exactly what you hear in Persian. Return your response as a JSON
		object with this exact format: \{"transcription": "your Persian transcription
		here"\}. Provide ONLY the JSON object, nothing else. Be accurate and precise in
		your transcription.
	\end{promptbox}
	Classification prompts always enumerate the permitted labels; the
	multiple-choice tasks list the options and require a single integer. Decoding is
	deterministic throughout: every model was called with
	\texttt{temperature}$=0.0$ and \texttt{max\_tokens}$=8192$.
	
	\subsection{Metrics}
	
	Classification tasks report macro-averaged F1 over the union of gold
	and predicted labels; multiple-choice tasks report accuracy; ASR reports WER and
	CER; translation reports COMET. Responses from which no valid label can be
	extracted, including refusals, are excluded from the scored set rather than
	counted as errors (Section~\ref{sec:asr}).
	
	\section{Experiments}
	\label{sec:experiments}
	
	\subsection{Models}
	\label{sec:models}
	
	We evaluate eight LALMs (Table~\ref{tab:models}, Appendix~\ref{app:supp}). We
	included every LALM with documented Persian audio support that was publicly
	available or API-accessible at submission time; models without documented
	Persian support, such as Phi-4-multimodal, were excluded.
	
	\subsection{Zero-shot results}
	
	\begin{table*}[t]
		\centering\footnotesize
		\setlength{\tabcolsep}{3pt}
		\renewcommand{\arraystretch}{1.05}
		\begin{tabular}{@{}lcc ccccc ccc@{}}
			\toprule
			\textbf{Task} & \textbf{Metric} & \textbf{Chance}
			& \multicolumn{5}{c}{\textbf{Open-weight}}
			& \multicolumn{3}{c}{\textbf{Proprietary} ($n{=}100$)} \\
			\cmidrule(lr){4-8}\cmidrule(l){9-11}
			& & & Q3-30B & Q2.5-7B & Q2.5-3B & G-E4B & G-E2B & Gem-2.5 & 4o & 4o-mini \\
			\midrule
			\multicolumn{11}{@{}l}{\textit{\textbf{Speech Understanding}}}\\
			ASR                     & WER md.$\downarrow$ & -- & \best{0.31} & 1.00 & 1.00 & 0.50 & 0.64 & 0.40 & 0.33 & \na \\
			Translation En$\to$Fa   & COMET & -- & 0.82 & 0.72 & 0.64 & 0.71 & 0.64 & \best{0.84} & 0.38 & 0.30 \\
			Translation Fa$\to$En   & COMET & -- & 0.60 & 0.46 & 0.44 & 0.68 & 0.64 & \best{0.82} & 0.50 & 0.48 \\
			Intent Detection        & F1    & 0.03 & 0.46 & 0.12 & 0.10 & 0.37 & 0.24 & \best{0.79} & 0.54 & 0.47 \\
			Named Entity Recog.\ & F1-ex    & --   & 0.14 & 0.01 & 0.01 & 0.14 & 0.09 & \best{0.45} & 0.24 & 0.17 \\
			Register (form./inf.) & F1    & 0.50 & \cellcolor{okaycell}0.74 & \cellcolor{badcell}0.38 & 0.59 & 0.59 & \cellcolor{badcell}0.49 & \best{0.88} & 0.77 & 0.68 \\
			Code-Switching          & F1    & 0.50 & \cellcolor{goodcell}\best{0.93} & \cellcolor{goodcell}0.82 & \cellcolor{badcell}0.42 & \cellcolor{badcell}0.51 & \cellcolor{badcell}0.37 & 0.92 & 0.28 & 0.35 \\
			Multiple-Choice QA      & Acc   & 0.25 & 0.40 & 0.28 & 0.27 & 0.33 & 0.28 & \best{0.64} & 0.42 & 0.40 \\
			Reading Comp.\ (Wiki)   & Acc   & 0.25 & \cellcolor{goodcell}0.92 & 0.79 & 0.60 & \cellcolor{goodcell}0.86 & \cellcolor{goodcell}0.85 & 0.87 & \best{0.94} & 0.91 \\
			Reading Comp.\ (Tiny)   & Acc   & 0.25 & 0.78 & 0.66 & 0.62 & 0.78 & 0.73 & 0.76 & 0.79 & \best{0.80} \\
			\midrule
			\multicolumn{11}{@{}l}{\textit{\textbf{Paralinguistic Analysis}}}\\
			Age Recognition     & F1 & 0.17 & \cellcolor{badcell}0.07 & \cellcolor{badcell}0.09 & \cellcolor{badcell}0.03 & \cellcolor{badcell}0.09 & \cellcolor{badcell}0.06 & \cellcolor{badcell}0.15 & \cellcolor{badcell}0.16 & \cellcolor{badcell}\best{0.19} \\
			Gender Recognition  & F1 & 0.50 & \cellcolor{goodcell}\best{0.99} & \cellcolor{goodcell}0.97 & \cellcolor{goodcell}0.97 & 0.60 & \cellcolor{badcell}0.41 & 0.82 & 0.60 & \cellcolor{badcell}0.43 \\
			Emotion Recognition & F1 & 0.18 & \best{0.56} & 0.41 & 0.38 & 0.27 & 0.20 & 0.48 & 0.36 & 0.30 \\
			\midrule
			\multicolumn{11}{@{}l}{\textit{\textbf{Persian Cultural Audio Understanding}}}\\
			Poetry Metre (\vazn) & Acc & 0.12 & \best{0.20} & \cellcolor{badcell}0.13& \cellcolor{badcell}0.10 & \cellcolor{badcell}0.11 & \cellcolor{badcell}0.12 & \cellcolor{badcell}0.10 & \cellcolor{badcell}0.10 & \cellcolor{badcell}0.12 \\
			Poetry Style (\sabk)& Acc & 0.64 & \cellcolor{badcell}\best{0.64} & \cellcolor{badcell}0.64 & \cellcolor{badcell}0.60 & \cellcolor{badcell}0.43 & \cellcolor{badcell}0.38 & \cellcolor{badcell}0.64 & \cellcolor{badcell}0.63 & \cellcolor{badcell}0.46 \\
			Music Understanding & Acc & 0.25 & \best{0.45} & 0.36 & 0.33 & 0.32 & 0.37 & 0.41 & 0.34 & 0.34 \\
			\bottomrule
		\end{tabular}
		\caption{Zero-shot audio performance. WER is lower-is-better; all other metrics
			are higher-is-better. WER and CER are fractions (0.36 denotes a 36\% error
			rate); refusals are excluded before scoring (Section~\ref{sec:asr}).
			\best{Bold} marks the best model per task. Shading:
			\colorbox{goodcell}{well above chance}, \colorbox{okaycell}{moderate},
			\colorbox{badcell}{at or below chance}. Chance is the
			majority-class share of the label set ($1/|\mathcal{L}|$ when
			balanced). Poetry-style accuracies are inflated by class imbalance:
			always predicting the majority class already scores 0.64. The ASR
			row reports median WER; means, refusal rates and
			repair analysis are in Table~\ref{tab:asr}. Confidence intervals are
			in Table~\ref{tab:cis}.
			\na: GPT-4o-mini refused all ASR items and has no score.}
		\label{tab:main}
	\end{table*}
	
	Table~\ref{tab:main} reports zero-shot audio performance. The overall picture
	follows the benchmark's three dimensions: models handle lexically driven
	speech-understanding tasks such as reading comprehension and code-switching
	reasonably well, do noticeably worse on pragmatic classification such as
	register, and struggle most with culturally grounded audio, where the
	relevant knowledge cannot be recovered from the words alone.

	Individual models depart from this picture in instructive ways.
	Qwen3-Omni-30B is the strongest open-weight model, with the best Persian ASR
	and the lead on most speech-understanding tasks, while the proprietary
	models, Gemini-2.5-Flash in particular, are ahead on translation and intent
	detection. Closed weights confer no advantage on cultural audio: poetry
	metre defeats proprietary and open models alike, with the sole exception of
	the largest open model (Section~\ref{sec:culture}).

	We report bootstrap confidence intervals for all headline results in
	Appendix~\ref{app:cis} and refer to them only where they affect the
	interpretation.
	
	\section{Analysis}
	\label{sec:analysis}
	
	\subsection{ASR error analysis}
	\label{sec:asr}
	
	\begin{table*}[h]
		\centering\footnotesize
		\setlength{\tabcolsep}{2.2pt}
		\begin{tabular}{@{}lrrrrrr@{}}
			\toprule
			\textbf{Model} & \textbf{Ref.} & \textbf{WER} & \textbf{WER} & \textbf{CER} & \textbf{Rep.} & \textbf{WER} \\
			& \textbf{\%}   & \textbf{mean}  & \textbf{med.}  & \textbf{mean}  & \textbf{\%}   & \textbf{fix} \\
			\midrule
			Qwen3-30B & 0.0 & \best{0.36} & \best{0.31} & \best{0.14} & 0.0 & 0.36 \\
			GPT-4o & 10.0 & 0.38 & 0.33 & 0.16 & 0.0 & 0.38 \\
			Gemini2.5F & 0.0 & 0.43 & 0.40 & 0.54 & 0.0 & 0.44 \\
			Qwen2.5-7B & 0.0 & 2.32 & 1.00 & 1.83 & 0.6 & 1.01 \\
			Qwen2.5-3B & 0.4 & 4.20 & 1.00 & 4.44 & 2.0 & 0.93 \\
			Gemma-E2B & 1.8 & 5.86 & 0.64 & 4.81 & 11.6 & 0.62 \\
			Gemma-E4B & 1.0 & 8.89 & 0.50 & 7.13 & 13.4 & \best{0.58} \\
			GPT-4o-mini & \textcolor{pbRed}{100} & \emph{ref.} & \emph{ref.} & \emph{ref.} & \emph{ref.} & \emph{ref.} \\
			\midrule
			\multicolumn{7}{@{}l@{}}{\textit{Reference systems}}\\
			SeamlessM4T-v2-lg & 0.0 & 0.24 & 0.13 & 0.11 & 0.0 & -- \\
			Whisper-fa-v4 & 0.0 & 0.29 & 0.20 & 0.13 & 0.0 & -- \\
			Whisper-lg-v3 & 0.0 & 0.30 & 0.29 & 0.10 & 0.0 & -- \\
			\bottomrule
		\end{tabular}
		\caption{ASR on PARSA-Bench. Ref.\% is the share of responses with no Persian
			script (refusals), excluded from scoring; \emph{ref.}\ marks a model that refused
			every item. Rep.\% is the share of scored utterances that are degenerate
			repetition loops; WER$^{\rm fix}$ is the mean WER after collapsing immediately
			repeated $n$-grams ($n \le 30$). Reference systems: SeamlessM4T-v2-large
			\citep{seamless2023}, Whisper-large-v3-Persian \citep{whisperpersian2024}
			and Whisper-large-v3 \citep{radford2023whisper}.}
		\label{tab:asr}
	\end{table*}
	
	Table~\ref{tab:asr} reports Persian ASR. Qwen3-Omni-30B transcribes Persian
	far better than any other evaluated LALM, evidence that large-scale
	multilingual pretraining transfers to Persian, while the smaller models
	degrade sharply -- the Gemma models are an order of magnitude worse. Even
	so, dedicated speech recognisers still lead every LALM on the same items by
	a clear margin, so general-purpose audio models approach but do not yet
	match specialised Persian ASR.

	Outputs from which no valid label can be extracted are excluded from the
	scored set rather than counted as errors, and the number of scored items
	accompanies every affected metric (Appendix~\ref{app:cis}). Exclusions are
	rare for the open-weight models but not for the proprietary ones: GPT-4o
	occasionally refuses audio-grounded requests, a behaviour also reported by
	\citet{Wang2024AudioBench}, and GPT-4o-mini declined the ASR task outright
	while answering the comprehension tasks normally, so it has no ASR score.
	Because refusals are excluded, the reported figures likely flatter these
	models' true transcription ability.

	The extreme error rates of the weaker models come from repetition collapse:
	instead of degrading smoothly, these models restate the same clause until
	they reach the token limit, so the hypothesis grows far longer than the
	reference. Collapses account for most of the mean WER in both Gemma-3n
	models, and collapsing repeated $n$-grams before scoring brings them within
	range of the proprietary systems. Qwen3-Omni-30B and the three proprietary
	models produce no collapses. Appendix~\ref{app:hallucination} gives the
	full breakdown.
	
	\subsection{The Role of ASR Quality in Downstream Performance}
	\label{sec:werqa}

\paragraph{Model-level analysis.}
We first ask whether models with stronger Persian ASR also perform better on downstream tasks. Table~\ref{tab:werall} (Appendix~\ref{app:supp}) compares each model's standalone ASR performance with its score on every benchmark task. The pattern is clear: models with lower WER consistently achieve better results on tasks whose answers depend primarily on lexical content, including translation, reading comprehension, intent detection, and named entity recognition. In contrast, this trend is much weaker for paralinguistic and culturally grounded tasks, where accurate transcription alone is often insufficient. Poetry metre is the main exception, as the only model performing above chance also achieves the best ASR performance.

\paragraph{Task-level analysis.}
We next examine whether transcription errors affect the same model differently across individual examples. For every evaluated item, we compare the model's transcription error with its downstream prediction on that item (Table~\ref{tab:iteml}, Appendix~\ref{app:supp}). The split runs along the same line as before. On content-transfer tasks, performance deteriorates steadily as transcription quality worsens: recognition errors propagate directly into the answer. On multiple-choice reasoning, paralinguistic analysis, and culturally grounded tasks, transcription quality has little effect item by item -- failures there arise from limitations beyond speech recognition.

The audio+text experiments (Section~\ref{sec:audiotext}) reinforce this interpretation. Providing the reference transcript substantially improves tasks that depend on recovering lexical content, particularly for examples with poor ASR. By contrast, it offers little benefit for tasks driven primarily by prosodic or other non-lexical cues.

	\subsection{The audio--text gap}
	\label{sec:gap}
	
	\begin{table}[t]
		\centering\footnotesize
		\setlength{\tabcolsep}{2.4pt}
		\begin{tabular}{@{}lccr@{}}
			\toprule
			\textbf{Task} & \textbf{Audio} & \textbf{Text} & \textbf{Gap} \\
			\midrule
			Trans.\ En$\to$Fa (COMET) & 0.824 & 0.856 & $-0.032$ \\
			Code-Switching (F1)       & 0.932 & 0.982 & $-0.050$ \\
			RC Wikipedia (Acc)        & 0.922 & 0.978 & $-0.056$ \\
			RC TinyStories (Acc)      & 0.780 & 0.920 & $-0.140$ \\
			Register (F1)             & 0.735 & 0.883 & $-0.148$ \\
			MCQA (Acc)                & 0.402 & 0.594 & $-0.192$ \\
			NER (F1-exact)            & 0.136 & 0.368 & $-0.232$ \\
			Trans.\ Fa$\to$En (COMET) & 0.595 & 0.830 & $-0.235$ \\
			Intent (F1)               & 0.461 & 0.787 & $-0.326$ \\
			\midrule
			Poetry Style (Acc)        & 0.640 & 0.541 & $+0.099$ \\
			Poetry Metre (Acc)        & 0.196 & 0.102 & $+0.094$ \\
			\bottomrule
		\end{tabular}
		\caption{Zero-shot audio versus text-only performance for
		Qwen3-Omni-30B, Gap $=$ audio $-$ text.}
		\label{tab:gap}
	\end{table}
	
	A central analysis in PARSA-Bench is the performance gap between zero-shot audio
	and text-only settings, which measures how much each task is affected by the
	additional challenge of processing speech. Table~\ref{tab:gap} shows that this
	gap is highly task dependent. Reading comprehension, code-switching, and
	En$\to$Fa translation show only minor degradation, since models can
	recover the relevant content even from imperfect speech representations. In
	contrast, intent detection, Fa$\to$En translation, and NER experience the
	largest drops; these tasks require reproducing the exact words of the
	input, so transcription errors propagate directly into the answer. The
	poetry tasks are the only cases where audio consistently beats text. For
	metre detection this is expected: the rhythmic pattern lives in the
	recitation and is not recoverable from unvowelled Persian orthography.
	Style classification shows the same direction across the Qwen family
	(Table~\ref{tab:sabk}), although performance in both conditions remains
	close to the majority baseline, which limits how much can be read into the
	difference (Section~\ref{sec:culture}).

\subsection{The Effect of Providing Reference Text Alongside Audio}
\label{sec:audiotext}

To complete the audio--text comparison, we evaluate a third zero-shot setting in
which models receive the original audio prompt together with the reference text
(transcript, source sentence, or passage) on the fourteen comparable tasks. The
results (Appendix~\ref{app:audiotext}) show that the benefit of combining
modalities is highly task dependent.

For tasks driven primarily by linguistic content, adding reference text largely
recovers the performance lost in the audio-only setting, bringing results close
to text-only performance in most cases. The largest improvements are observed
for tasks such as code-switching and translation, where accurate lexical
information is essential. Register detection shows the clearest evidence of
multimodal synergy, with models benefiting from the combination of lexical and
prosodic cues.

Poetry style behaves differently: adding the transcript consistently
reduces performance across models, consistent with stylistic information in
poetry recitation being carried mainly by acoustic features rather than the
words themselves. For paralinguistic tasks, where transcripts provide little
task-relevant information, the additional text has minimal impact, neither
substantially improving nor degrading performance.

	\subsection{Few-shot and chain-of-thought prompting}
	\label{sec:fewshot}
	
	{
	Table~\ref{tab:fscot} (Appendix~\ref{app:supp}) reports Qwen3-Omni-30B under
	few-shot and chain-of-thought prompting. Chain-of-thought mainly benefits
	extraction-style tasks such as intent detection, where working through the
	options helps, but it often hurts tasks that rely on direct perception
	rather than explicit reasoning, including reading comprehension, emotion
	recognition, and code-switching, where the reasoning trace adds noise to a
	judgement that needs none. Few-shot audio demonstrations produce mixed
	effects -- mild gains on tasks with large answer spaces, losses on the
	binary ones -- and combining the two is volatile rather than additive. No
	prompting variant closes the gap to the text-only condition, which remains
	the ceiling on every task where it exists.
	}
	
\subsection{Paralinguistics and cultural understanding}
\label{sec:culture}

Paralinguistic tasks reveal substantial variation across capabilities. Gender
recognition is mostly reliable for the Qwen models across scales, while
Gemma-3n-E2B performs near chance, so the ability appears to require a
certain level of model capacity. Emotion recognition is harder: the strongest
models beat the random baseline but stay far from ceiling, leaving
fine-grained affective perception in Persian speech an open problem. Age
recognition is the most difficult of the three -- every model performs close
to chance, consistent with how little age information the voice carries
without visual or contextual cues.

Poetry metre remains particularly challenging because it depends on
recognising rhythmic patterns of syllable weight that Persian orthography
does not write down. Only the largest model shows evidence of this
capability, and even there it is limited: its correct answers are
concentrated on a small subset of metres rather than spread across the label
set, while every other model, open or proprietary, performs at chance. Metre
perception, in other words, is not merely weak in current models -- it is
absent below the largest scale tested, and narrow even there.

Poetry style presents a different problem: performance is dominated by the
majority class. Because the style set is naturally imbalanced, always
answering the most frequent category already matches the best model's
accuracy, and the low macro-F1 confirms that minority styles are rarely
identified. On this formulation, no model reliably distinguishes stylistic
categories. Music understanding is weak in a revealing way: the culturally
central skill, Dastgah identification, is at chance for every model, and the
best scores come from shallower acoustic properties such as polyphony
detection.
	
	\subsection{Impact of model scale}
	\label{sec:scale}
	
	\begin{figure}[t]
		\centering
		\includegraphics[width=0.98\columnwidth]{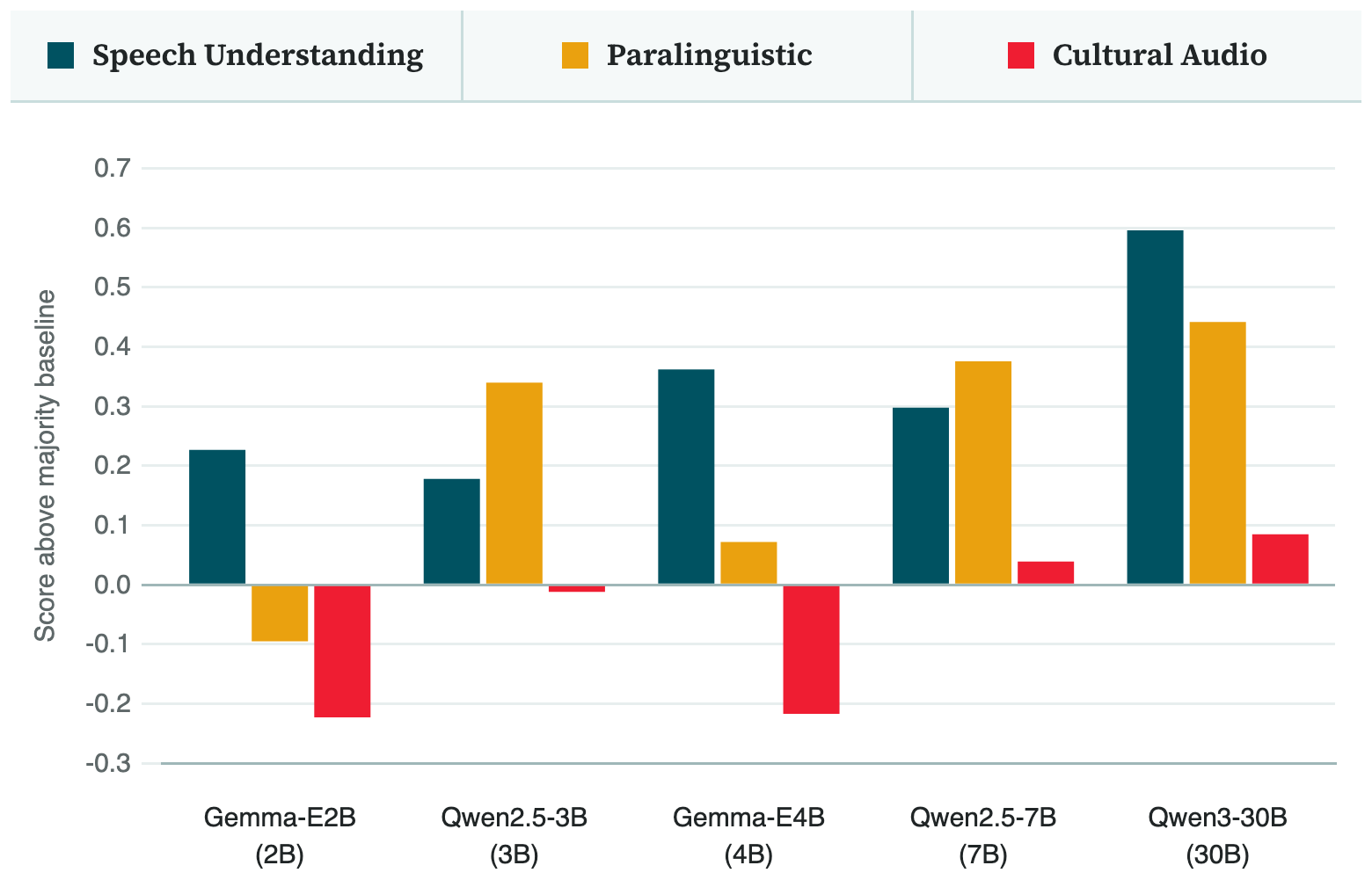}
		\caption{Average score per evaluation dimension for the open-weight models,
			ordered by parameter count. Each bar is the macro-average of
			$(\mathrm{score}-b)/(1-b)$ over the classification tasks in the
			dimension, where $b$ is the task's majority-class baseline, so
			$0$ means majority-level and negative bars are below it; ASR and
			translation are excluded because WER and COMET have no chance
			level. Scale
			improves performance consistently within the Qwen family, but Gemma-3n-E4B (4B)
			outperforms Qwen2.5-Omni-7B on Speech Understanding despite fewer parameters,
			showing that training-data coverage and encoder quality matter alongside model
			size.}
		\label{fig:scale}
	\end{figure}
	
	Scale alone is not a reliable predictor of performance, echoing findings
	from multilingual text benchmarks. Figure~\ref{fig:scale} plots each
	open-weight model's average score above the majority baseline in each of
	the three dimensions.

	Performance improves steadily with scale within the Qwen family, but the
	trend does not carry across architectures: the smaller Gemma-3n-E4B beats
	the larger Qwen2.5-Omni-7B on several speech-understanding tasks, among
	them Persian-to-English translation and Wikipedia reading comprehension.
	Where clean transcription matters, encoder quality counts for more than
	decoder size; on culturally grounded and pragmatic tasks, the Qwen family
	draws on broader multilingual pretraining, likely including Persian text,
	an advantage that scale within the Gemma family cannot replicate. The
	smallest Gemma model collapses on the paralinguistic tasks altogether.
	Scale helps, but only when paired with domain-relevant data and a strong
	audio encoder.
	
	\section{Conclusion}
	\label{sec:conclusion}
	
We introduced PARSA-Bench, a benchmark for evaluating Large Audio-Language
Models on Persian speech, language, and cultural understanding across 16
tasks. Our evaluation of eight LALMs points to audio processing, not
linguistic knowledge, as the primary obstacle: text-only models outperform
their audio counterparts on most tasks with matched conditions, and supplying
the reference transcript alongside the audio recovers much of the lost
performance. The models, in other words, often know what they need to know
but cannot reliably extract it from speech.

The picture varies sharply by task type. Lexically grounded tasks are
increasingly within reach of stronger models, while paralinguistic tasks such
as age and emotion recognition remain difficult at every scale. The cultural
tasks expose the deepest gaps. Poetry style classification benefits from
acoustic information unavailable in text, although no model exceeds the style
majority-class baseline (Section~\ref{sec:culture}); poetry metre remains
largely unsolved, with only the largest evaluated model showing above-chance
performance. Prosody cuts both ways here: tasks that depend on it gain from
audio alone and can even be degraded by additional text.

Neither scale nor closed weights guarantees cultural audio understanding --
proprietary flagship models do not consistently beat smaller open-weight ones
on culturally grounded tasks, so specialised data and training strategies
likely matter more than model size.

These findings highlight several directions for future work, including the
development of Persian prosodic speech resources for metre-aware modelling,
retrieval-based approaches for culturally grounded knowledge, Persian-specific
audio representation learning, and broader evaluation of low-resource languages
with rich cultural audio traditions.
	
	\section*{Limitations}
	\label{sec:limitations}
	
	While PARSA-Bench covers a broad range of Persian audio phenomena, several
	limitations should be noted. First, six tasks rely on TTS-synthesised audio,
	which may underrepresent the prosodic variability and disfluency of natural
	speech; future work could complement these with fully naturalistic recordings.
	Second, the benchmark does not yet include human performance baselines, which
	would provide a meaningful upper bound, particularly for culturally grounded
	tasks such as \vazn{} detection where near-chance model performance invites
	comparison to human judgements. Third, due to the API costs associated with
	proprietary models, extended prompting experiments, including few-shot and
	chain-of-thought conditions, were conducted only for open-weight models; a
	fuller prompting analysis across Gemini and GPT-4o remains an avenue for future
	work. Fourth, all models were evaluated in zero-shot and prompting-only regimes;
	fine-tuned or retrieval-augmented systems may yield substantially different
	results, especially on culturally specific tasks. Finally, PARSA-Bench is
	intentionally scoped to Persian; extending this evaluation framework to other
	low-resource languages with rich oral traditions remains an important direction
	for future work.

	{
	Several construct-validity caveats should also be noted. In the
	code-switching and register tasks, label is fully aligned with source
	(all positive code-switching items come from YouTube and all negative
	items from Common Voice; all formal register items from Mana-TTS and all
	informal items from GPTInformal-Persian), so channel or synthesis
	artefacts could contribute to performance independently of the intended
	construct; a channel-only control is left to future work. In the poetry
	tasks, recitations come from 58 reciters, 46 of whom appear in exactly one
	style class, so reciter identity is partially confounded with style.
	Music clips are uniformly segmented to about five seconds, which may be
	too short to establish a Dastgah, and the at-chance Dastgah result should
	be read with this in mind. Because most source corpora are public,
	text-only baselines may additionally be inflated by pretraining
	memorisation; the TinyStories set, whose passages are novel by
	construction, partially controls for this. Finally, the
	reading-comprehension distractors were generated with GPT-4o-mini, and
	GPT-4o models are among the systems evaluated on them, so a
	self-preference effect cannot be excluded.
	}

	{
	\section*{Ethics Statement}

	\paragraph{Data provenance and licences.} Common Voice and CoVoST2 are
	released under CC0, ParsiNLU and TinyStories under permissive research
	licences, and SHEMO for research use. The code-switching subset includes
	audio extracted from public Persian YouTube channels; we redistribute
	short excerpts for non-commercial research evaluation together with the
	source video identifiers, so that provenance is auditable and removal
	requests can be honoured, and we do not redistribute full videos. Ganjoor
	recitations are used with attribution to the hosting library and their
	reciter identifiers are preserved. PARSA-Bench itself is released for
	non-commercial research use.

	\paragraph{Synthetic voices.} TTS audio was generated with a Persian TTS
	system using reference voices from Common Voice, whose contributors
	released their recordings under CC0 for arbitrary reuse including
	synthesis. No voice is presented as belonging to an identified individual,
	and we release only the resulting evaluation audio, not voice-cloning
	capability.

	\paragraph{Demographic inference.} Age and gender labels are self-reported
	by Common Voice contributors and are used solely to evaluate model
	behaviour. Gender is operationalised as the binary distinction available
	in the source metadata, which does not represent the full range of gender
	identities; this is a limitation inherited from the corpus. Voice-based
	demographic inference is dual-use, and we discourage using these subsets
	to build demographic-inference systems, which carry well-known
	discrimination risks.

	\paragraph{Intended use.} PARSA-Bench is an evaluation resource, not a
	training set, and results on it should not be treated as evidence of
	deployment readiness for Persian speech applications.
	}

	\emergencystretch=2em
	\bibliography{parsa}
	
	\appendix
	
	\section{Evaluated models and supplementary results}
	\label{app:supp}
	
	\begin{table}[h]
		\centering\footnotesize
		\setlength{\tabcolsep}{4pt}
		\begin{tabular}{@{}lccl@{}}
			\toprule
			\textbf{Model} & \textbf{Par.} & \textbf{Type} & \textbf{Ref.} \\
			\midrule
			Qwen2.5-Omni-3B     & 3B  & Open  & \citep{Qwen2025Omni} \\
			Qwen2.5-Omni-7B     & 7B  & Open  & \citep{Qwen2025Omni} \\
			Qwen3-Omni-30B-A3B & 30B & Open & \citep{Qwen2025Omni} \\
			Gemma-3n-E2B        & 2B  & Open  & \citep{GoogleGemma3} \\
			Gemma-3n-E4B        & 4B  & Open  & \citep{GoogleGemma3} \\
			GPT-4o-mini-audio & -- & Prop. & \citep{OpenAI2024GPT4o} \\
			GPT-4o-audio        & --  & Prop. & \citep{OpenAI2024GPT4o} \\
			Gemini-2.5-Flash    & --  & Prop. & \citep{GoogleGemini25} \\
			\bottomrule
		\end{tabular}
		\caption{Evaluated LALMs. All accept audio natively and generate Persian text.}
		\label{tab:models}
	\end{table}
	
	\begin{table}[h]
		\centering\footnotesize
		\setlength{\tabcolsep}{2.6pt}
		\begin{tabular}{@{}lr@{\hskip 8pt}lr@{}}
			\toprule
			\multicolumn{2}{c}{\textit{Lexically driven}} & \multicolumn{2}{c}{\textit{Paraling.\ / cultural}} \\
			\cmidrule(r){1-2}\cmidrule(l){3-4}
			Intent (F1)        & $-0.99$ & Metre (Acc) & $-0.66$ \\
			RC Tiny (Acc)      & $-0.96$ & Music (Acc) & $-0.58$ \\
			RC Wiki (Acc) & $-0.83$ & Age (F1)        & $-0.31$ \\
			MCQA (Acc)   & $-0.76$ & Emotion (F1)    & $-0.17$ \\
			Register (F1)      & $-0.72$ & Gender (F1)     & $+0.28$ \\
			Code-Switch.\ (F1) & $-0.28$ & Style (Acc)     & $+0.36$ \\
			\midrule
			\textbf{mean} & $\mathbf{-0.75}$ & \textbf{mean} & $\mathbf{-0.18}$ \\
			\bottomrule
		\end{tabular}
		\caption{Correlation between a model's \emph{standalone} Persian ASR median WER
			and its score on each task, across the five open-weight models. Negative means
			better transcription goes with a better score. With five models per correlation
			these are indicative, not significance tests.}
		\label{tab:werall}
	\end{table}

{
\begin{table}[h]
	\centering\footnotesize
	\setlength{\tabcolsep}{2.6pt}
	\begin{tabular}{@{}lrrrrr@{}}
		\toprule
		\textbf{Task} & \textbf{$r$} & \textbf{Gain$_{Q1}$} & \textbf{Gain$_{Q4}$} & \textbf{Fix$_{\rm hi}$} & \textbf{Fix$_{\rm lo}$} \\
		\midrule
		Fa$\to$En (chrF)   & $-0.29$ & $+0.24$ & $+0.32$ & --   & --   \\
		En$\to$Fa (chrF)   & $-0.22$ & $+0.08$ & $+0.13$ & --   & --   \\
		Intent             & $-0.15$ & $+0.22$ & $+0.42$ & 0.48 & 0.42 \\
		NER (F1)           & $-0.15$ & $+0.18$ & $+0.30$ & --   & --   \\
		RC Wiki            & $-0.11$ & $+0.09$ & $+0.18$ & 0.88 & 0.84 \\
		\midrule
		MCQA               & $-0.04$ & $+0.10$ & $+0.12$ & 0.32 & 0.31 \\
		RC Tiny            & $-0.04$ & $+0.13$ & $+0.13$ & 0.65 & 0.65 \\
		Register           & $-0.04$ & $+0.22$ & $+0.25$ & 0.77 & 0.75 \\
		Emotion            & $-0.04$ & $+0.00$ & $+0.05$ & 0.26 & 0.25 \\
		Gender             & $+0.02$ & $+0.00$ & $-0.03$ & 0.14 & 0.20 \\
		Metre              & $+0.02$ & $+0.00$ & $+0.00$ & 0.03 & 0.04 \\
		Age                & $+0.04$ & $+0.08$ & $+0.03$ & 0.18 & 0.17 \\
		Style              & $+0.05$ & $-0.03$ & $-0.07$ & 0.11 & 0.17 \\
		Code-Switch.       & $+0.07$ & $+0.17$ & $+0.12$ & 0.63 & 0.60 \\
		\bottomrule
	\end{tabular}
	\caption{Item-level analysis, averaged over the five open-weight models.
		For each item, the model's own transcription WER on that item's audio is
		joined to the same model's zero-shot and audio+text outcomes on the same
		item. $r$ = Pearson correlation between per-item WER (capped at 1.0) and
		the zero-shot outcome; Gain$_{Q1}$/Gain$_{Q4}$ = audio+text minus audio
		outcome on the best- and worst-transcribed WER quartiles;
		Fix$_{\rm hi}$/Fix$_{\rm lo}$ = share of zero-shot errors repaired by the
		supplied transcript among items above/below the model's median WER
		(binary and multiple-choice tasks only). Tasks above the rule show the
		content-transfer regime; tasks below show no item-level dependence.}
	\label{tab:iteml}
\end{table}
}

{
\begin{table}[h]
	\centering\footnotesize
	\setlength{\tabcolsep}{3pt}
	\begin{tabular}{@{}lccccc@{}}
		\toprule
		\textbf{Task} & \textbf{ZS} & \textbf{FS} & \textbf{CoT} & \textbf{CoT-FS} \\
		\midrule
		MCQA (Acc)     & 0.402 & 0.404 & 0.392 & 0.532 \\
		RC Wiki (Acc)  & 0.922 & 0.926 & 0.758 & 0.904 \\
		RC Tiny (Acc)  & 0.784 & 0.802 & 0.718 & 0.782 \\
		Intent (F1)    & 0.461 & 0.518 & 0.694 & 0.723 \\
		Code-Sw.\ (F1) & 0.932 & 0.864 & 0.820 & --    \\
		Register (F1)  & 0.735 & 0.657 & 0.723 & 0.468 \\
		Emotion (F1)   & 0.559 & 0.486 & 0.393 & 0.419 \\
		\bottomrule
	\end{tabular}
	\caption{Qwen3-Omni-30B across prompting conditions: zero-shot (ZS),
	few-shot audio (FS), chain-of-thought (CoT) and their combination.
	``--'' marks a condition that was not run.}
	\label{tab:fscot}
\end{table}

}
	
	\begin{table}[h]
		\centering\footnotesize
		\setlength{\tabcolsep}{3pt}
		\begin{tabular}{@{}lccc@{}}
			\toprule
			\textbf{Model} & \textbf{Audio} & \textbf{Text} & \textbf{A$-$T} \\
			\midrule
			Qwen2.5-7B & 0.636 & 0.615 & $+0.021$ \\
			Qwen3-30B & \best{0.640} & 0.541 & $+0.099$ \\
			Qwen2.5-3B & 0.597 & 0.600 & $-0.003$ \\
			Gemma-E4B & 0.425 & 0.525 & $-0.100$ \\
			Gemma-E2B & 0.385 & 0.529 & $-0.144$ \\
			\bottomrule
		\end{tabular}
		\caption{Poetry style (\sabk): zero-shot audio versus text-only accuracy,
			$n=442$, four options, majority-class share $=0.64$.}
		\label{tab:sabk}
	\end{table}

{
\section{Audio-plus-text evaluation}
\label{app:audiotext}

Table~\ref{tab:audiotext} reports the complete
zero-shot audio-plus-text condition (Section~\ref{sec:audiotext}): each model
receives the unchanged zero-shot audio prompt, the audio, and the reference
text of the audio. A+T is the audio-plus-text score with a percentile
bootstrap 95\% confidence interval; A and T are the same model's audio-only
and text-only baselines; $\Delta =$ A+T $-\max($A, T$)$. Style baselines are paired audio-only
controls run with an identical prompt.
Text-only runs for age, gender and emotion do not exist because the
transcript does not carry those labels; their $\Delta$ is against audio
alone. Fa$\to$En few-shot audio has COMET $0.547$ for
Qwen3-Omni-30B, against $0.595$ zero-shot.

\begin{table*}[p]
	\centering\footnotesize
	\setlength{\tabcolsep}{2.4pt}
	\begin{minipage}[t]{0.49\textwidth}
		\centering
		\begin{tabular}{@{}lcccr@{}}
			\toprule
		\textbf{Model} & \textbf{A+T [95\% CI]} & \textbf{A} & \textbf{T} & \textbf{$\Delta$} \\
\midrule
\multicolumn{5}{@{}l}{\textit{Intent (F1)}}\\
Qwen3-30B & 0.720 [0.645,0.739] & 0.460 & 0.787 & $-0.067$ \\
Q2.5-7B & 0.551 [0.481,0.573] & 0.120 & 0.615 & $-0.063$ \\
Q2.5-3B & 0.444 [0.372,0.463] & 0.100 & 0.560 & $-0.115$ \\
E4B & 0.586 [0.523,0.605] & 0.370 & 0.722 & $-0.137$ \\
E2B & 0.513 [0.448,0.534] & 0.240 & 0.668 & $-0.156$ \\
\midrule
\multicolumn{5}{@{}l}{\textit{NER (F1)}}\\
Qwen3-30B & 0.342 [0.306,0.379] & 0.140 & 0.386 & $-0.044$ \\
Q2.5-7B & 0.263 [0.230,0.297] & 0.010 & 0.325 & $-0.062$ \\
Q2.5-3B & 0.249 [0.218,0.280] & 0.010 & 0.257 & $-0.007$ \\
E4B & 0.349 [0.313,0.386] & 0.140 & 0.379 & $-0.029$ \\
E2B & 0.330 [0.300,0.362] & 0.090 & 0.374 & $-0.044$ \\
\midrule
\multicolumn{5}{@{}l}{\textit{Register (F1)}}\\
Qwen3-30B & 0.859 [0.824,0.887] & 0.740 & 0.883 & $-0.025$ \\
Q2.5-7B & 0.825 [0.789,0.856] & 0.380 & 0.549 & $+0.275$ \\
Q2.5-3B & 0.790 [0.752,0.824] & 0.590 & 0.621 & $+0.170$ \\
E4B & 0.869 [0.835,0.896] & 0.590 & 0.830 & $+0.039$ \\
E2B & 0.855 [0.822,0.883] & 0.490 & 0.850 & $+0.004$ \\
\midrule
\multicolumn{5}{@{}l}{\textit{Code-Sw.\ (F1)}}\\
Qwen3-30B & 0.984 [0.972,0.994] & 0.930 & 0.982 & $+0.002$ \\
Q2.5-7B & 0.887 [0.860,0.915] & 0.820 & 0.832 & $+0.055$ \\
Q2.5-3B & 0.962 [0.944,0.978] & 0.420 & 0.333 & $+0.542$ \\
E4B & 0.759 [0.721,0.798] & 0.510 & 0.827 & $-0.068$ \\
E2B & 0.413 [0.379,0.450] & 0.370 & 0.692 & $-0.279$ \\
\midrule
\multicolumn{5}{@{}l}{\textit{MCQA (Acc)}}\\
Qwen3-30B & 0.596 [0.552,0.638] & 0.400 & 0.594 & $+0.002$ \\
Q2.5-7B & 0.396 [0.354,0.436] & 0.280 & 0.438 & $-0.042$ \\
Q2.5-3B & 0.290 [0.250,0.328] & 0.270 & 0.354 & $-0.064$ \\
E4B & 0.430 [0.385,0.470] & 0.330 & 0.434 & $-0.004$ \\
E2B & 0.350 [0.312,0.392] & 0.248 & 0.371 & $-0.021$ \\
\midrule
\multicolumn{5}{@{}l}{\textit{RC Wiki (Acc)}}\\
Qwen3-30B & 0.986 [0.976,0.994] & 0.920 & 0.978 & $+0.008$ \\
Q2.5-7B & 0.958 [0.938,0.976] & 0.790 & 0.970 & $-0.012$ \\
Q2.5-3B & 0.858 [0.824,0.890] & 0.600 & 0.930 & $-0.072$ \\
E4B & 0.982 [0.970,0.992] & 0.860 & 0.968 & $+0.014$ \\
E2B & 0.980 [0.968,0.990] & 0.850 & 0.972 & $+0.008$ \\
\midrule
\multicolumn{5}{@{}l}{\textit{RC Tiny (Acc)}}\\
Qwen3-30B & 0.922 [0.898,0.944] & 0.780 & 0.920 & $+0.002$ \\
Q2.5-7B & 0.876 [0.844,0.902] & 0.660 & 0.872 & $+0.004$ \\
Q2.5-3B & 0.724 [0.684,0.760] & 0.620 & 0.768 & $-0.044$ \\
E4B & 0.890 [0.860,0.916] & 0.780 & 0.900 & $-0.010$ \\
E2B & 0.860 [0.830,0.888] & 0.730 & 0.888 & $-0.028$ \\
		\bottomrule
		\end{tabular}
	\end{minipage}\hfill
	\begin{minipage}[t]{0.49\textwidth}
		\centering
		\begin{tabular}{@{}lcccr@{}}
			\toprule
		\textbf{Model} & \textbf{A+T [95\% CI]} & \textbf{A} & \textbf{T} & \textbf{$\Delta$} \\
\midrule
\multicolumn{5}{@{}l}{\textit{Age (F1)}}\\
Qwen3-30B & 0.125 [0.098,0.151] & 0.070 & -- & $+0.055$ \\
Q2.5-7B & 0.000 [0.000,0.000] & 0.090 & -- & $-0.090$ \\
Q2.5-3B & 0.142 [0.117,0.165] & 0.030 & -- & $+0.112$ \\
E4B & 0.119 [0.094,0.145] & 0.090 & -- & $+0.029$ \\
E2B & 0.105 [0.080,0.128] & 0.060 & -- & $+0.045$ \\
\midrule
\multicolumn{5}{@{}l}{\textit{Gender (F1)}}\\
Qwen3-30B & 0.984 [0.974,0.994] & 0.990 & -- & $-0.006$ \\
Q2.5-7B & 0.958 [0.940,0.976] & 0.970 & -- & $-0.012$ \\
Q2.5-3B & 0.938 [0.916,0.958] & 0.970 & -- & $-0.032$ \\
E4B & 0.581 [0.538,0.625] & 0.600 & -- & $-0.019$ \\
E2B & 0.470 [0.427,0.508] & 0.410 & -- & $+0.060$ \\
\midrule
\multicolumn{5}{@{}l}{\textit{Emotion (F1)}}\\
Qwen3-30B & 0.537 [0.485,0.585] & 0.560 & -- & $-0.023$ \\
Q2.5-7B & 0.380 [0.331,0.420] & 0.410 & -- & $-0.030$ \\
Q2.5-3B & 0.314 [0.279,0.351] & 0.380 & -- & $-0.066$ \\
E4B & 0.394 [0.344,0.440] & 0.270 & -- & $+0.124$ \\
E2B & 0.345 [0.293,0.392] & 0.200 & -- & $+0.145$ \\
\midrule
\multicolumn{5}{@{}l}{\textit{Metre (Acc)}}\\
Qwen3-30B & 0.222 [0.179,0.268] & 0.196 & 0.102 & $+0.026$ \\
Q2.5-7B & 0.118 [0.084,0.150] & 0.127 & 0.153 & $-0.035$ \\
Q2.5-3B & 0.101 [0.072,0.133] & 0.101 & 0.117 & $-0.016$ \\
E4B & 0.133 [0.098,0.167] & 0.112 & 0.109 & $+0.021$ \\
E2B & 0.115 [0.084,0.150] & 0.118 & 0.102 & $-0.003$ \\
\midrule
\multicolumn{5}{@{}l}{\textit{Style (Acc)}}\\
Qwen3-30B & 0.584 [0.536,0.636] & 0.636 & 0.541 & $-0.052$ \\
Q2.5-7B & 0.584 [0.541,0.629] & 0.643 & 0.615 & $-0.059$ \\
Q2.5-3B & 0.437 [0.389,0.484] & 0.518 & 0.600 & $-0.163$ \\
E4B & 0.595 [0.550,0.643] & 0.609 & 0.525 & $-0.014$ \\
E2B & 0.631 [0.588,0.681] & 0.606 & 0.529 & $+0.025$ \\
\midrule
\multicolumn{5}{@{}l}{\textit{Fa$\to$En (COMET)}}\\
Qwen3-30B & 0.831 [0.819,0.842] & 0.600 & 0.830 & $+0.000$ \\
Q2.5-7B & 0.795 [0.782,0.806] & 0.460 & 0.794 & $+0.001$ \\
Q2.5-3B & 0.757 [0.744,0.770] & 0.440 & 0.762 & $-0.005$ \\
E4B & 0.842 [0.831,0.852] & 0.680 & 0.835 & $+0.007$ \\
E2B & 0.833 [0.822,0.843] & 0.640 & 0.801 & $+0.032$ \\
\midrule
\multicolumn{5}{@{}l}{\textit{En$\to$Fa (COMET)}}\\
Qwen3-30B & 0.852 [0.841,0.862] & 0.820 & 0.856 & $-0.004$ \\
Q2.5-7B & 0.752 [0.738,0.767] & 0.720 & 0.738 & $+0.014$ \\
Q2.5-3B & 0.652 [0.637,0.667] & 0.640 & 0.657 & $-0.005$ \\
E4B & 0.854 [0.844,0.863] & 0.710 & 0.751 & $+0.103$ \\
E2B & 0.839 [0.828,0.849] & 0.640 & 0.719 & $+0.120$ \\
		\bottomrule
		\end{tabular}
	\end{minipage}
	\caption{Audio-plus-text results for all fourteen tasks and the five
	open-weight models, zero-shot. Left: speech-understanding tasks; right:
	paralinguistic, cultural and translation tasks. Age/Qwen2.5-7B scored on
	only 4 of 480 items (output-format collapse) and is reported for
	completeness.}
	\label{tab:audiotext}
\end{table*}
}

	\section{Prompts}
	\label{app:prompts}
	
	All prompts are reproduced verbatim from the released evaluation code. Braces
	shown as \texttt{\{\dots\}} are Python format fields filled at run time.
	
	\begin{promptbox}[{ASR (all models)}]
		You are an expert Persian speech recognition system. Listen to the Persian audio
		and transcribe exactly what you hear in Persian. Return your response as a JSON
		object with this exact format: \{"transcription": "your Persian transcription
		here"\}. Provide ONLY the JSON object, nothing else. Be accurate and precise in
		your transcription.
	\end{promptbox}
	
	\begin{promptbox}[{Speech translation, En$\to$Fa}]
		You are an expert English-to-Persian translator. You MUST process the provided
		audio input directly. Never claim you cannot process audio. Listen to the
		English audio and translate it to Persian. Return your response as a JSON object
		with this exact format: \{"translation": "your Persian translation here"\}.
		Provide ONLY the JSON object, nothing else. Be accurate and precise in your
		translation.
	\end{promptbox}
	The Fa$\to$En prompt is identical with the languages exchanged.
	
	\begin{promptbox}[{Intent detection}]
		Task: Persian Speech-to-Intent Classification\\
		Instruction: Listen carefully to the Persian audio and classify the user's
		intent. You must choose the most appropriate intent from the list below.\\[2pt]
		Allowed Intents: \{intent\_list\}\\[2pt]
		Constraint: Your response must be a single JSON object. Do not include any
		conversational text, explanations, or transcriptions. Use the exact keys and
		values provided.\\[2pt]
		Output Format: \{"intent": "<intent\_name>"\}\\
		Classification:
	\end{promptbox}
	
	\begin{promptbox}[{Named entity recognition}]
		Task: Persian Speech Named Entity Recognition (NER)\\
		Instruction: Listen carefully to the Persian audio and identify all named
		entities. Extract entities and classify them by type.\\[2pt]
		Allowed Entity Types: \{entity\_types\}\\[2pt]
		Constraint: Your response must be a single JSON object. Do not include any
		conversational text, explanations, or transcriptions.\\[2pt]
		Output format:\\
		\{"entities": [\{"type": "entity\_type", "value": "entity\_value"\}, ...]\}\\
		Entities (JSON):
	\end{promptbox}
	
	\begin{promptbox}[{Code-switching detection}]
		You are an expert in Persian language analysis. Your task is to detect if a
		Persian audio clip contains code-switching which is mixing Persian with
		English.\\[2pt]
		Respond with ONLY 'True' if code-switching happens, or 'False' if it is not.
	\end{promptbox}
	
	\begin{promptbox}[{Formal/informal register}]
		You are an expert in Persian language analysis. Your task is to classify spoken
		Persian speech as either FORMAL or INFORMAL based on linguistic characteristics,
		on the vocabulary, grammar, tone, and overall structure of the speech.\\[2pt]
		Respond with ONLY with one word 'FORMAL' or 'INFORMAL'.
	\end{promptbox}
	
	\begin{promptbox}[{Multiple-choice QA}]
		Task: Persian Multiple Choice Question Answering\\
		Instruction: Listen to the Persian question and answer the multiple-choice
		question.\\[2pt]
		Steps:\\
		1. Listen carefully to the question\\
		2. Evaluate each option (A, B, C, D)\\
		3. Select the single best answer\\[2pt]
		\#\#\# Question Audio:\\
		\#\#\# Options:\\
		\{options\}\\[2pt]
		Constraint: Your response must be a single JSON object containing only the
		answer.\\
		Output format: \{"answer": "A"\} (where the value is one of: A, B, C, or D)\\
		Answer (JSON):
	\end{promptbox}
	
	\begin{promptbox}[{Reading comprehension}]
		Task: Persian Reading Comprehension - Multiple Choice\\
		Instruction: Listen to the Persian audio passage and answer the multiple-choice
		question based on what you heard.\\[2pt]
		Steps:\\
		1. Listen carefully to the entire audio passage\\
		2. Read the question\\
		3. Evaluate each option (A, B, C, D)\\
		4. Select the single best answer\\[2pt]
		\#\#\# Question:\\\{question\}\\
		\#\#\# Options:\\\{options\}\\[2pt]
		Constraint: Your response must be a single JSON object containing only the
		answer.\\
		Output format: \{"answer": "A"\} (where the value is one of: A, B, C, or D)\\
		Answer (JSON):
	\end{promptbox}
	
	\begin{promptbox}[{Emotion recognition}]
		Task: Persian Speech Emotion Recognition\\
		Instruction: Listen carefully to the Persian speech and classify the speaker's
		emotion.\\[2pt]
		Constraints:\\
		- Choose exactly one label that best describes the emotion from: anger,
		happiness, sadness, neutral, surprise, fear\\
		- Do NOT explain your reasoning\\
		- Output JSON only in this format: \{"emotion": "<chosen\_label>"\}\\
		Emotion (JSON):
	\end{promptbox}
	
	\begin{promptbox}[{Age recognition}]
		You are an expert in Persian speech analysis. Your task is to identify the
		speaker's age group.\\
		IMPORTANT INSTRUCTIONS:\\
		- You must respond with ONLY one of these options: \{age\_map\}\\
		- Provide ONLY the integer of the option.\\
		- Do not include extra text or reasoning.
	\end{promptbox}
	where \texttt{\{age\_map\}} enumerates the six Common Voice brackets.
	
	\begin{promptbox}[{Gender recognition}]
		You are a gender classification system. Your task is to identify the speaker's
		gender by carefully listening to the vocal characteristics of the Persian audio,
		including pitch, tone, resonance, and speaking style.\\
		IMPORTANT INSTRUCTIONS:\\
		- You must respond with ONLY male or female\\
		- Do not include the any extra text\\
		- Do not add any explanation or reasoning
	\end{promptbox}
	
	\begin{promptbox}[{Poetry metre (\vazn)}]
		You are an expert in Persian poetry Vazn classification. Your task is to
		identify the correct poetic Vazn by listening to its audio that is provided for
		you.\\
		Options:\\
		\{ten\_options\}\\
		IMPORTANT INSTRUCTIONS:\\
		- Listen carefully to the rhyme scheme and rhythmic meter.\\
		- Respond with ONLY the integer from 1 to 10 of the correct option.\\
		- Do not include any extra text or reasoning.
	\end{promptbox}
	
	\begin{promptbox}[{Music understanding}]
		You are a Persian music expert tasked with answering multiple-choice music
		questions. Listen carefully to the audio and analyze its musical aspects
		(instruments, melody, rhythm, dastgah, genre, tempo, etc.).\\[2pt]
		Question: \{question\}\\
		Options:\\\{choices\}\\[2pt]
		IMPORTANT INSTRUCTIONS:\\
		- Select the best answer from the provided choices.\\
		- Respond with ONLY the integer (1, 2, 3, or 4) of the correct option.\\
		- Do not include any extra text or reasoning.\\
		- Base your answer strictly on Persian music knowledge.
	\end{promptbox}
	
	\paragraph{Few-shot and CoT variants.} Few-shot prompts prepend $k=2$
	demonstrations. CoT prompts extend the JSON schema with a \texttt{reasoning}
	field emitted before the answer field.
	
	\paragraph{Dataset construction: reading-comprehension options.}
	GPT-4o-mini, \texttt{temperature}$=0.8$, one call per source question:
	\begin{promptbox}
		You are an expert psychometrician specializing in Persian reading comprehension
		MCQ creation.\\[2pt]
		Your task: Convert an existing question-answer pair into a multiple-choice
		question with 4 options (A, B, C, D).\\[2pt]
		CRITICAL REQUIREMENTS:\\
		1. Keep the EXACT original question text - do NOT modify it\\
		2. Use the provided answer as the CORRECT option\\
		3. Create 3 plausible DISTRACTORS that:\\
		\ \ - Are factually incorrect based on the context\\
		\ \ - Use similar vocabulary/style to the correct answer\\
		\ \ - Are similar in length to the correct answer\\
		\ \ - Sound convincing to someone who didn't read carefully\\
		4. RANDOMIZE which position (A/B/C/D) contains the correct answer\\
		5. Use formal Persian\\
		6. Assess question DIFFICULTY: easy / medium / hard\\[2pt]
		Return ONLY valid JSON (no markdown, no extra text): \{"question": \dots,
		"options": \{"A": \dots, "D": \dots\}, "correct\_answer": "A/B/C/D",
		"difficulty": \dots, "explanation": \dots\}
	\end{promptbox}
	The TinyStories set uses the same model at \texttt{temperature}$=0.9$ with a
	per-question-type instruction instead of a source question-answer pair. Option
	order is additionally shuffled in post-processing.
	
	\paragraph{Dataset construction: music-question paraphrasing.}
	GPT-4o-mini, \texttt{temperature}$=0.7$:
	\begin{promptbox}
		You are an expert in Persian language paraphrasing.\\[2pt]
		Your task: Paraphrase the given question while maintaining the EXACT same
		meaning.\\[2pt]
		REQUIREMENTS:\\
		1. Keep the meaning IDENTICAL - don't change what is being asked\\
		2. Use different words and sentence structure\\
		3. Maintain formal Persian\\
		4. Keep the same question type (what, when, who, why, how, etc.)\\
		5. The paraphrased question should be natural and fluent\\[2pt]
		Return ONLY valid JSON: \{"paraphrased\_question": \dots\}
	\end{promptbox}
	
	\section{Audio duration and label statistics}
	\label{app:duration}
	
	\begin{table}[h]
		\centering\footnotesize
		\setlength{\tabcolsep}{3.5pt}
		\begin{tabular}{@{}lrrrrr@{}}
			\toprule
			\textbf{Task} & \textbf{$n$} & \textbf{mean} & \textbf{min} & \textbf{max} & \textbf{total} \\
			&              & (s)           & (s)          & (s)          & (h) \\
			\midrule
			ASR                     & 500 & 4.92 & 1.42 & 20.12 & 0.68 \\
			Translation Fa$\to$En   & 500 & 5.35 & 2.42 & 10.54 & 0.74 \\
			Translation En$\to$Fa   & 499 & 5.67 & 1.94 & 21.98 & 0.79 \\
			Intent Detection        & 500 & 3.03 & 0.98 &  6.94 & 0.42 \\
			Named Entity Rec.\      & 497 & 3.26 & 1.02 &  9.64 & 0.45 \\
			Register                & 496 & 4.21 & 0.24 & 18.04 & 0.58 \\
			Code-Switching          & 500 & 6.55 & 1.51 & 17.68 & 0.91 \\
			MCQA                    & 500 & 6.29 & 1.24 & 31.88 & 0.87 \\
			Age Recog.\            & 480 & 3.61 & 1.91 &  6.23 & -- \\
			Gender Recog.\         & 500 & 2.45 & 0.62 &  6.39 & -- \\
			Emotion Recognition     & 388 & 3.57 & 0.41 & 23.40 & 0.39 \\
			Poetry Metre            & 347 & 15.12 & 5.70 & 49.33 & 1.46 \\
			Poetry Style            & 442 & 29.17 & 13.91 & 49.05 & 3.58 \\
			Music Underst.\         & 496 & 4.94 & 0.61 &  5.31 & 0.68 \\
			RC Wikipedia            & 500 & 28.68 & 2.66 & 40.00 & 3.98 \\
			RC TinyStories          & 500 & 36.57 & 7.08 & 40.00 & 5.08 \\
			\midrule
			\textbf{Total}  & \textbf{7{,}645} & & & & \textbf{20.6} \\
			\bottomrule
		\end{tabular}
		\caption{Audio duration statistics of the evaluated subsets.}
		\label{tab:durations}
	\end{table}

	\section{Repetition collapse in ASR}
	\label{app:hallucination}
	
	\begin{table}[h]
		\centering\footnotesize
		\setlength{\tabcolsep}{2.6pt}
		\begin{tabular}{@{}lrrrrr@{}}
			\toprule
			\textbf{Model} & \textbf{Rep.\%} & \textbf{WER}$_{\rm rep}$ & \textbf{WER}$_{\rm ok}$ & \textbf{share} & \textbf{WER}$^{\rm fix}$ \\
			\midrule
			Qwen3-30B    & 0.0  & --    & 0.36 & 0\%  & 0.36 \\
			Gemini-2.5F  & 0.0  & --    & 0.43 & 0\%  & 0.44 \\
			GPT-4o       & 0.0  & --    & 0.56 & 0\%  & 0.56 \\
			Qwen2.5-7B   & 0.6  & 58.7  & 1.98 & 15\% & 1.01 \\
			Qwen2.5-3B   & 2.0  & 98.0  & 2.27 & 47\% & 0.93 \\
			Gemma-E2B    & 11.6 & 44.6  & 1.32 & 82\% & 0.62 \\
			Gemma-E4B    & 13.4 & 57.0  & 1.45 & 86\% & 0.58 \\
			\bottomrule
		\end{tabular}
		\caption{Repetition collapse on the ASR task.
			Rep.\% = share of utterances flagged as collapses;
			WER$_{\rm rep}$ and WER$_{\rm ok}$ = mean WER on the collapsed
			and non-collapsed subsets; share = fraction of the model's mean WER
			contributed by the collapsed subset; WER$^{\rm fix}$ = mean WER over
			all utterances after collapsing repeated $n$-grams. Values in this table are
			computed before refusal exclusion, and GPT-4o-mini is omitted because it
			refused every item (Section~\ref{sec:asr}).}
		\label{tab:repetition}
	\end{table}
	
	Table~\ref{tab:repetition} gives the full picture behind Section~\ref{sec:asr}.
	An utterance is flagged as a collapse when the hypothesis is at least $3\times$
	the reference length and one token occupies at least $30\%$ of it. The failure
	is confined to the four weakest open-weight models: Qwen3-Omni-30B,
	Gemini-2.5-Flash and GPT-4o never produce a collapse. For the two Gemma models
	the collapsed subset, one utterance in eight, contributes over $80\%$ of the
	reported mean WER, so the headline number describes a decoding-termination
	failure rather than transcription quality. This also explains the divergence
	between mean and median for the weak models.
	
	\section{Confidence intervals}
	\label{app:cis}
	
	\begin{table}[h]
		\centering\footnotesize
		\setlength{\tabcolsep}{2.6pt}
		\begin{tabular}{@{}llcr@{}}
			\toprule
			\textbf{Task} & \textbf{Model} & \textbf{Score [95\% CI]} & \textbf{$n$} \\
			\midrule
			\multirow{5}{*}{Code-Sw.\ (F1)}
			& Qwen3-30B   & 0.932 [0.908, 0.952] & 500 \\
			& Qwen2.5-7B  & 0.819 [0.785, 0.851] & 500 \\
			& Qwen2.5-3B  & 0.423 [0.386, 0.459] & 500 \\
			& Gemma-E4B   & 0.511 [0.467, 0.553] & 500 \\
			& Gemma-E2B   & 0.368 [0.340, 0.398] & 500 \\
			\midrule
			\multirow{5}{*}{Register (F1)}
			& Qwen3-30B   & 0.735 [0.693, 0.774] & 496 \\
			& Qwen2.5-7B  & 0.375 [0.341, 0.412] & 496 \\
			& Qwen2.5-3B  & 0.590 [0.543, 0.635] & 496 \\
			& Gemma-E4B   & 0.588 [0.543, 0.633] & 496 \\
			& Gemma-E2B   & 0.488 [0.445, 0.533] & 496 \\
			\midrule
			\multirow{5}{*}{RC Wiki (Acc)}
			& Qwen3-30B   & 0.922 [0.898, 0.944] & 500 \\
			& Qwen2.5-7B  & 0.788 [0.754, 0.822] & 500 \\
			& Qwen2.5-3B  & 0.600 [0.556, 0.644] & 500 \\
			& Gemma-E4B   & 0.864 [0.836, 0.894] & 500 \\
			& Gemma-E2B   & 0.848 [0.816, 0.880] & 500 \\
			\midrule
			\multirow{5}{*}{Emotion (F1)}
			& Qwen3-30B   & 0.559 [0.503, 0.609] & 388 \\
			& Qwen2.5-7B  & 0.413 [0.370, 0.457] & 381 \\
			& Qwen2.5-3B  & 0.376 [0.335, 0.417] & 388 \\
			& Gemma-E4B   & 0.268 [0.225, 0.312] & 388 \\
			& Gemma-E2B   & 0.203 [0.162, 0.245] & 388 \\
			\midrule
			\multirow{5}{*}{Metre (Acc)}
			& Qwen3-30B   & 0.196 [0.156, 0.239] & 347 \\
			& Qwen2.5-7B  & 0.127 [0.092, 0.161] & 347 \\
			& Qwen2.5-3B  & 0.101 [0.072, 0.133] & 347 \\
			& Gemma-E4B   & 0.112 [0.081, 0.147] & 347 \\
			& Gemma-E2B   & 0.118 [0.086, 0.153] & 347 \\
			\midrule
			\multirow{5}{*}{Style (Acc)}
			& Qwen3-30B   & 0.640 [0.595, 0.688] & 442 \\
			& Qwen2.5-7B  & 0.636 [0.593, 0.686] & 442 \\
			& Qwen2.5-3B  & 0.597 [0.550, 0.647] & 442 \\
			& Gemma-E4B   & 0.425 [0.378, 0.473] & 442 \\
			& Gemma-E2B   & 0.385 [0.337, 0.432] & 442 \\
			\midrule
			\multirow{5}{*}{Music (Acc)}
			& Qwen3-30B   & 0.452 [0.409, 0.496] & 496 \\
			& Qwen2.5-7B  & 0.363 [0.319, 0.403] & 496 \\
			& Qwen2.5-3B  & 0.329 [0.288, 0.371] & 496 \\
			& Gemma-E4B   & 0.321 [0.284, 0.363] & 496 \\
			& Gemma-E2B   & 0.371 [0.329, 0.413] & 496 \\
			\midrule
			\multirow{3}{*}{ASR (WER$\downarrow$)}
			& Qwen3-30B   & 0.358 [0.334, 0.383] & 500 \\
			& Qwen2.5-7B  & 2.317 [1.397, 3.556] & 500 \\
			& Gemma-E4B   & 8.895 [6.915, 10.963] & 500 \\
			\bottomrule
		\end{tabular}
		\caption{Percentile bootstrap 95\% confidence intervals.}
		\label{tab:cis}
	\end{table}
	
\end{document}